\def\year{2022}\relax
\documentclass[letterpaper]{article} 
\usepackage{aaai22}  
\usepackage{times}  
\usepackage{helvet}  
\usepackage{courier}  
\usepackage[hyphens]{url}  
\usepackage{graphicx} 
\usepackage{natbib}  
\usepackage{caption} 
\DeclareCaptionStyle{ruled}{labelfont=normalfont,labelsep=colon,strut=off} 
\usepackage{algorithm}
\usepackage{algorithmic}
\usepackage{amsmath}
\usepackage{bbm}
\usepackage{amssymb}
\usepackage{newfloat}
\usepackage{listings}
\floatstyle{ruled}
\newfloat{listing}{tb}{lst}{}
\floatname{listing}{Listing}
\usepackage{graphicx}
\usepackage{graphicx}
\usepackage{color}

\title{SimCLAD: A Simple Framework for Contrastive Learning of \\ Acronym Disambiguation}
\author{
    Bin Li\textsuperscript{\rm 1},
    Fei Xia\textsuperscript{\rm 2,3},
    Yixuan Weng\textsuperscript{\rm 2},
    Xiusheng Huang\textsuperscript{\rm 2,3},
   	Bin Sun\textsuperscript{\rm 1}\thanks{Corresponding Author.}
}
\affiliations{
    \textsuperscript{\rm 1} College of Electrical and Information Engineering, Hunan University\\
    \textsuperscript{\rm 2} National Laboratory of Pattern Recognition, Institute of Automation, Chinese Academy Sciences\\
    \textsuperscript{\rm 3} School of Artificial Intelligence, University of Chinese Academy of Sciences\\

	\{libincn, sunbin611\}@hnu.edu.cn 
	\\
	\{xiafei2020, huangxiusheng2020\}@ia.ac.cn, \text wengsyx@gmail.com
}

\begin{document}

\maketitle

\begin{abstract}
Acronym disambiguation means finding the correct meaning of an ambiguous acronym from the dictionary in a given sentence, which is one of the key points for scientific document understanding (SDU@AAAI-22). Recently, many attempts have tried to solve this problem via fine-tuning the pre-trained masked language models (MLMs) in order to obtain a better acronym representation. {However, the acronym meaning is varied under different contexts, whose corresponding phrase representation mapped in different directions lacks discrimination in the entire vector space.} Thus, the original representations of the pre-trained MLMs are not ideal for the acronym disambiguation task. In this paper, we propose a \textbf{Sim}ple framework for \textbf{C}ontrastive \textbf{L}earning of \textbf{A}cronym \textbf{D}isambiguation (\textbf{SimCLAD}) method to better understand the acronym meanings. Specifically, we design a continual contrastive pre-training method that enhances the pre-trained model's generalization ability {by learning the phrase-level contrastive distributions between true meaning and ambiguous phrases.} The results on the acronym disambiguation of the scientific domain in English show that the proposed method outperforms all other competitive state-of-the-art (SOTA) methods.
\end{abstract}
\section{Introduction}
Recently, the pre-training technology has highly improved the machine understanding level \cite{qiu2020pre}. However, due to the complexity and ambiguity of the natural language, there is still a gap between the machines and humans in comprehensively understanding  documents \cite{veyseh2020acronym}. In scientific document understanding (SDU@AAAI-22), due to space limitations, the appearance of acronyms is often necessary. It is of great significance to correctly understand and distinguish the correct acronym meaning from the given sentence \cite{veyseh-et-al-2022-Multilingual}.
\par
More precisely, the document reading system is expected to find the correct expanded form of the acronym given the possible expansions from the dictionary for the acronym.  This is quite important for a variety of downstream tasks containing the understanding part, such as reading comprehension \cite{gardner2019making}, story cloze \cite{guan2021lot} and medical entity disambiguation \cite{li2021more}, etc.
\par
\begin{figure}[t]
	\centering
	\includegraphics[scale=0.50]{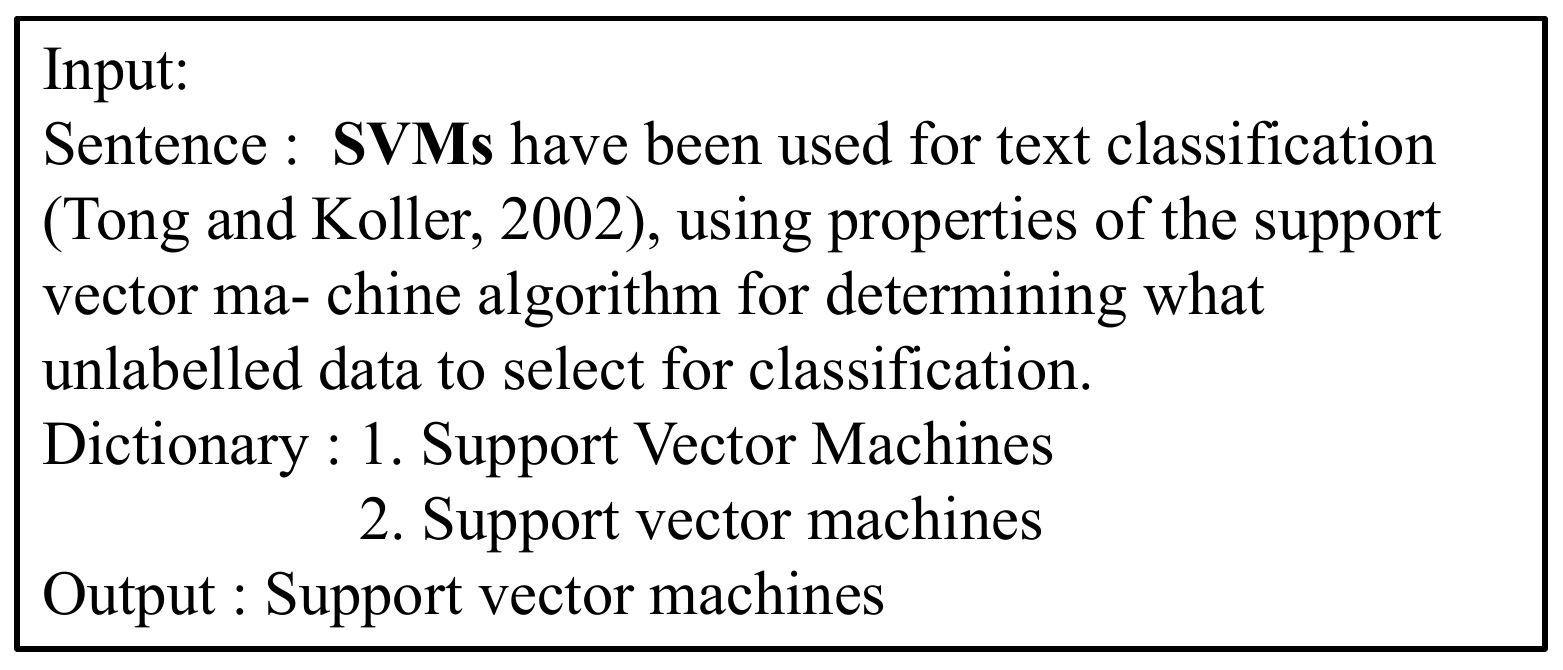}
	\caption{Example of acronym disambiguation.}
	\label{fig1}
	\vspace{-0.5cm}
\end{figure}

The acronym disambiguation task aims at finding the correct meaning of the ambiguous acronym in a given text from the dictionary \cite{veyseh2020does}. As shown in Figure \ref{fig1}, the sentence is from the scientific domain in English, where the text in bold represents the short acronym. The dictionary contains the indistinguishable acronym of long-form. Our goal is to predict the correct meaning of the long-form acronyms from the dictionary (i.e., Support vector machines). A good prediction should not only understand the context meaning, but also differ the meaning of ambiguous phrases. Many works have attempted to incorporate the manually designed rules \cite{schwartz2002simple}, handcrafted features \cite{luo2018attention, li2021systems}, word embedding \cite{jaber2021participation} and pre-training technology \cite{zhong2021leveraging, kubal2021effective, pan2021bert} into this task and achieved relatively good performance. According to the result of the SDU@AAAI-21 \cite{veyseh2020acronym}, the pre-training method can effectively outperform the rule-based or feature-based method by a large margin. However,
the acronym meaning varies in different contexts \cite{veyseh2021maddog}, {whose corresponding token representation is anisotropic distribution \cite{su2021tacl}. For the masked language models (MLMs), the token representation is mapped with a cramped idiomatic distribution \cite{bert, su2021tacl}.} As a result, the MLMs are weak in distinguishing the ambiguous meaning of acronyms, especially in the acronym disambiguation task.
\par
\par 
{Inspired by the token-aware contrastive learning method \cite{su2021tacl},} a \textbf{Sim}ple framework for \textbf{C}ontrastive \textbf{L}earning of \textbf{A}cronym \textbf{D}isambiguation   (\textbf{Sim-CLAD}) me-thod is proposed to distinguish the distributions between true meaning and ambiguous phrases. {Specifically, we adopt the phrase-level continual contrastive pre-training method to enhance the pre-trained MLMs for a better representation of the acronyms.} Extensive experiments carried on the acronym disambiguation of the scientific domain in English show that the proposed method achieves the best results compared with the other competitive state-of-the-art (SOTA) methods. The online leaderboard shows that the proposed method ranks 1-st in the scientific English domain in the shared task2 of the SDU@AAAI-22.
The main contributions are summarized as follows:
\begin{itemize}
\item We {perform} the first attempt to resolve the acronym disambiguation problem with a contrastive pre-trained model for better acronym understanding.
\item {We extend the token-level contrastive learning method by designing a phrase-level continuing contrastive pre-training method to obtain better contrastive representations of the ambiguous acronyms.}
\item Experiments conducted on the scientific English dataset demonstrate that the proposed method has better performance and outperforms other competitive baselines.
\end{itemize}
\vspace{-0.2cm}
\section{Related work}
\subsection{Acronym disambiguation}
Acronym disambiguation has attracted much attention in biomedical fields \cite{jin2019deep}.
The earliest methods \cite{schwartz2002simple} utilize manually designed rules or text features to find out the acronym expansions. Later, there have been a few works \cite{nadeau2005supervised} on automatically digging out the acronym expansions by analyzing the web data. 
These methods are usually effective when an acronym appears in conjunction with the corresponding extensions in the same document. 
However, traditional rules or statistics cannot effectively handle these tasks with the explosive growth of information. In addition, these methods used for biomedical tasks cannot be directly transferred to other fields, such as science.
Recently, deep learning based methods have promoted the development of scientific document understanding (SDU). Methods like feature-based \cite{luo2018attention}, clustering \cite{jaber2021participation}, and pre-training model methods \cite{pan2021bert} perform well in this task. Although these methods based on the pre-training technology (i.e., MLMs) can effectively distinguish confusing phrases of the acronym, they still lack the cognition of negative samples in the representation. Different from the above methods, we use contrast learning to obtain more obvious features for acronym disambiguation.
\subsection{Contrastive learning}
In general, methods based on contrastive learning (CL) can well distinguish the observed data from other negative samples. Many attempts of the CL have been made to many areas of computer vision, including image \cite{chopra2005learning} and video \cite{wang2015unsupervised}. Most recently, a simple framework for the CL of visual representations named as SimCLR \cite{chen2020simple} based on NT-Xent loss is proposed for better image representation. The same idea can also be found in the field of natural language processing (NLP). In the field of NLP, many works \cite{wu2020clear, liu2021fast} are devoted to modeling better sentence-level representations with the CL for the downstream tasks. {Recently, Su et al. \cite{su2021tacl} propose a token-aware CL framework to learn the isotropic and discriminative distribution of token representations by restoring the original token meaning of the masked items. This method is very effective in distinguishing the token-level representations thereby achieving better performance in sentence representation.} Following this work, we further consider the phrase-level CL by recovering the probable phrases (i.e., ambiguous acronyms) during the pre-training phase to obtain a better-distinguished acronym representation. 
\subsection{Continual pre-training}
It is a wise choice for further continual pre-training the pre-trained model \cite{gururangan2020don} to alleviate the task and domain discrepancy between the upstream and the downstream tasks. Many works tend to investigate how to better transform the general knowledge to the domain-specific task via continuing pre-training \cite{han2021econet, su2021tacl}. 
In the field of the SDU, the generic MLMs are weak in well distinguishing confusing phrases from the dictionary. As a result, the continual pre-training method is adopted in this paper to directly improve the ability of understanding with contrastive learning.
\section{Task introduction}
\subsection{Problem definition}
The acronym disambiguation task aims to find
the correct meaning of a given acronym in a sentence. Specifically, the sentence can be represented as $X = [x_1, x_2, \ldots, x_n]$, where $n$ is the total length of the sentence. Given that the index $i$ represents the acronym in the input sentence, the short acronym can be represented as $\hat{x}_{i}$, The corresponding meaning of the short-form acronym is chosen from the dictionary $D = [T_1, T_2, \ldots, T_k]$, where the $T_k$ represents the phrase in the dictionary, and the $k$ represents the total length of the probable phrases. Our goal is to predict the correct phrase meaning $S_j$ of short acronym $\hat{x}_{i}$ from the dictionary $D$, where the $j \in [1,k], i \in [1,n]$.
\subsection{Evaluation metric}
To evaluate the performance of different methods, the Macro F1 is adopted. The definitions are shown as follows:
\begin{equation}
\begin{aligned}
\text { Precision } &=\frac{\sum_{i=1}^{n} \text { precision}_{i}}{n} \\
\text { Recall } &=\frac{\sum_{i=1}^{n} \text { recall}_{i}}{n} \\
\text{Macro F}1 &=\frac{2 \times \text { Precision } \times \text { Recall }}{\text { Precision }+\text { Recall }}
\end{aligned}
\end{equation}
where $n$ is the number of total classes, the precision$_i$ and
recall$_i$ represent the precision and recall of class $i$ respectively.
\begin{figure*}[t]
	\centering
	\includegraphics[scale=0.61]{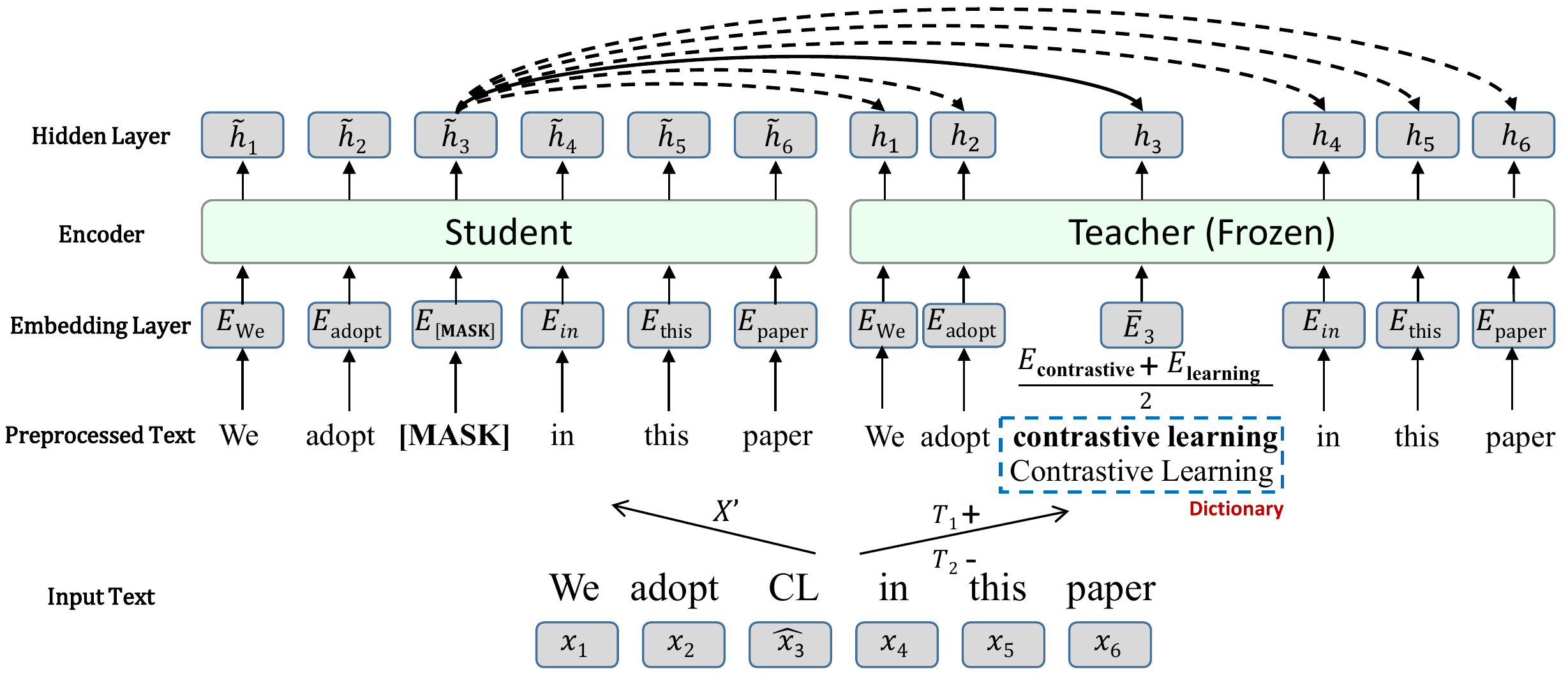}
	\caption{{Overview of the proposed method, where the student learns the masked acronym closer to its real meaning produced by the teacher (solid arrow) and away from the other confusing phrases in the dictionary (dashed arrows). The phrase embedding is averaged before encoding.} }
	\label{fig2}
	\vspace{-0.4cm}
\end{figure*}
\subsection{Dataset}
\begin{table}[h]
	\centering
	\renewcommand\arraystretch{1.2}
	\caption{Statistical information of scientific English dataset.}
	\begin{tabular}{c|cc}
			\noalign{\hrule height 1pt}
		\textbf{Data} & \textbf{Sample Number} & \textbf{Ratio} \\ 
			\noalign{\hrule height 0.5pt}
		\textbf{Training Set}       & 7532        & 83.69\%         \\ 
		\textbf{Development Set}      & 894        & 9.93\%         \\ 
		\textbf{Test Set}       & 574       & 6.38\%         \\ 			\noalign{\hrule height 0.5pt}
		\textbf{Total}      & 9000         & 100\%         \\ 			\noalign{\hrule height 1pt}
	\end{tabular}

	\label{table1}
			\vspace{-0.3cm}
\end{table}
		\vspace{-0.1cm}
The acronym disambiguation contains the dataset of scientific English, which is shown in Table \ref {table1}. The dataset is divided into training (7532), development (894), and testing (574) according to the data set. All the datasets can be found in the work \cite{veyseh-et-al-2022-MACRONYM}, where the training and validation sets of the scientific English dataset have been manually labeled. All the labels are collected in the dictionary.
\section{Method}
\subsubsection{Model architecture}
As shown in Figure 2, the overview of the proposed method contains two domain pre-trained models (a student and a teacher) which are initialized with the same parameters, i.e., SciBERT (Beltagy, Lo, and Cohan 2019). At the stage of pre-training, the parameters of the teacher are frozen to provide a good encoding representation for the student model. In addition, the teacher supports the well-formed original objectives of the MLM (i.e., masked language modeling and next sentence prediction) for the student model. {Inspired by \cite{su2021tacl}, we intentionally mask the original short-form acronym ($X^{'}$) to perform the distinguish ambiguous long-form acronyms ($T_1+, T_2-$) in the teacher model, where notation $+$ and $-$ are positive and negative samples.} A contrast loss is adopted in the pre-training process of the student model. Specifically, it is obtained by masking the short-form acronym (i.e., CL) in the input sentence of the student model against the ``correct meaning'' produced by the teacher without masking the corresponding phrases. To get the representation of the ``reference'' phrase in the dictionary (dotted frame), we perform phrase averaged method by averaging the embeddings of the tokens (i.e., contrastive learning), which is presented with the upper bar. Meanwhile, we let the representation distance of positive and negative (i.e., Contrastive Learning) samples stay away to enhance the model's ability to distinguish confusing samples.
\vspace{-0.1cm}
\subsubsection{Phrase-level contrastive pre-training}
The proposed me-thod is composed of two pre-trained models who are both initialized with the SciBERT model, where the one is the student model (noted as $S$) and the other is the teacher model (noted as $T$). During the pre-training phase, we only optimize the parameters of $S$ leaving the $T$ model to be frozen. {Given an input sentence $X = [x_1, \ldots, x_n]$, we intentionally mask the short acronym $\hat x_i$ following the same pre-training task \cite{bert}. Then, we feed the masked sentence $X^{'}$ into the student to perform the pre-trained training task. As a result, we obtain the contextual representation $\widetilde h = [\widetilde h_1, \ldots, \widetilde h_n]$ in the student model, where the $\textbf{[MASK]}$ is embedded as $E_{\textbf{[mask]}}$. At the same time, the teacher model replaces the corresponding short acronym $\hat{x}_{i}$ in the original sentence $X$ with the phrase in the dictionary $D$ as input. It is intuitive that the teacher can distinguish all the probable representations with the dictionary, where we want the student model to distinguish the correct phrase meaning through CL. In the end, the well-formed phrase representation is utilized with the averaged embeddings. Thus, the final representation of the recovered sentence $h = [h_1, \ldots, h_n]$ against the corresponding input sentence is produced by the teacher (see Figure \ref{fig2}). Following the work \cite{su2021tacl}, we further refine the proposed phrase-level contrastive pre-training loss}
\begin{equation}
\mathcal{L}_{\mathrm{CL}} = -\sum_{i=1}^{n}\sum_{k=1}^{K} \mathbbm{1}\left(\hat{x}_{i}, {T}_{k}\right) \log \frac{e^{ \text{S}\left(\widetilde{h}_{i}, h_{i}\right) / \tau}}{\sum_{j=1}^{n} e^{ \text{S}\left(\widetilde{h}_{i}, h_{j}\right) / \tau}},
\label{eq1}
\end{equation}
{where the indicator function $\mathbbm{1}(\hat{x}_{i}, {T}_{k}) = 1$ if $\hat x_i$ is the masked acronym and short for the corresponding long-form $T_k$. Otherwise, $\mathbbm{1}(\hat{x}_{i},  {T}_{k}) = 0$. We use the $\tau$ as the temperature hyper-parameter and the notation $\text{S}(·,·)$ represents the similarity function, where we choose the cosine function. The $K$ is the number of the all possible long-form acronyms.}
\subsubsection{Optimizing objectives}
Naturally, the student model lear-ns to distinguish the masked acronym closer to its corresponding ``true'' representation produced by the teacher and away from the meaning of the other confusing phrases in the sentence. {In summary, the acronym representations learned by the student are more discriminative with the confusing phrases, therefore better following an isotropic distribution \cite{su2021tacl}.}
Furthermore, the original pre-training method of the MLM \cite{bert} is also adopted for learning good document representations, including the masked language modeling task and the next sentence prediction (NSP) task. The overall optimizing objectives are performed as the continual pre-training in the domain-specific corpus, which can be shown as 
\begin{equation}
\mathcal{L}=\mathcal{L}_{\mathrm{CL}}+\mathcal{L}_{\mathrm{MLM}}+\mathcal{L}_{\mathrm{NSP}}
\end{equation}
where the pre-training step is totally unsupervised, which can be carried out with the vast scientific English dataset. Once the pre-trained model is obtained, the student model will be fine-tuned on the acronym disambiguation task.
\subsubsection{Contrastive fine-tuning}
Concretely, given the final hidden state, $h_{x}$ of the input sentence, the representation of the probable phrases can be represented as $h_{T_j}$. We concatenate the $h_{x}$ and the $h_{T_j}$ to obtain the feature $\boldsymbol {h}$ for the two classification and contrastive learning, which can be presented as 
\begin{equation}
\boldsymbol{h}=\left[{h_{x}; {h}_{{T_j}}}\right]
\end{equation}
A non-linear projection layer is added on top of the pre-trained model for obtaining  representation. The positive sample is noted as $+$, and the negative sample is noted as $-$. The calculation of two types of the feature can be shown as follows:
\begin{equation}
\begin{aligned}
&z_{i}^{+}=W_{2} \operatorname{ReLU}\left(W_{1} \boldsymbol{h}^{+}\right) \\
&z_{j}^{-}=W_{2} \operatorname{ReLU}\left(W_{1} \boldsymbol{h}^{-}\right)
\end{aligned}
\end{equation}
Finally, we perform fine-tuning in a multi-task manner and take a weighted average of the two classification losses and the contrastive loss:
\begin{equation}
\mathcal{L}=\frac{(1-\lambda)}{2}\left(\mathcal{L}_{C E}^{+}+\mathcal{L}_{C E}^{-}\right)+\lambda \mathcal{L}_{\mathrm{CL}}
\end{equation}
where the $\lambda$ is the weight hyper-parameter.

\section{Experiment setup}
\subsection{Baseline models}
\begin{itemize}
\item \textbf{Rule-based method} The baseline method proposed by Schwartz is a rule-based method \cite{schwartz2002simple}. In this baseline, the similarity of the candidate long-forms with the sample text (in terms of several overlapping words) is first computed. Then, the long-form with the highest similarity score is chosen as the final prediction. The related codes can be found on the website\footnote{https://github.com/amirveyseh/AAAI-22-SDU-shared-task-2-AD}.
\item \textbf{RoBERTa model}
The RoBERTa \cite{liu2019roberta} is mainly trained on general domain corpora with Byte Pair Encoding\cite{sennrich2016neural} based on the original structure of the BERT.
This model can provide a good fine-grained representation of the sentence which can be used in distinguishing acronyms.
\item \textbf{SciBERT model} The SciBERT \cite{beltagy2019scibert} is a domain-specific pre-trained language model for science. This architecture of the SciBERT follows the same architecture as BERT to capture the well-formed representation of the scientific data. This model has achieved better performance than the original BERT-based method in some scientific tasks, which can be viewed as a good backbone for the acronym disambiguation.
\item \textbf{hdBERT model} 
The hdBERT model \cite{zhong2021leveraging} considers the domain agnostic and specific knowledge, adopting the hierarchical dual-path BERT method jointly trained with fine-grained and high-level specific representations for acronym disambiguation. The context-based pre-trained models including the RoBERTa and the SciBERT are elaborately involved in encoding these two kinds of knowledge respectively. Finally, the multiple layer perception is devised to integrate to output the prediction.
\item \textbf{BERT-MT model}
The BERT-MT method \cite{pan2021bert} is designed with a binary classification model incorporating the BERT and several training strategies including dynamic negative sample selection, task adaptive pretraining, adversarial training, and pseudo labeling. This method achieves the best performance in the SDU@AAAI-21 competition of the scientific English, which is the strong baseline.
\end{itemize}
\begin{table*}[t]
	\centering
	\renewcommand\arraystretch{1.2}	\setlength{\tabcolsep}{4mm}
	\caption{F1 performance in scientific English.}
	\begin{tabular}{cccc}
		\noalign{\hrule height 1pt}
		Method     & Macro Precision  & Macro Recall  & \ \ Macro F1 \\ \noalign{\hrule height 0.5pt}
		Rule-based & 0.74       & 0.37        &\ \  0.49   \\ 
		RoBERTa & 0.81       & 0.78   &\ \  0.79   \\ 
		SciBERT  & 0.85      & 0.82       &\ \ 0.83    \\ 
		hdBERT   & 0.89       & 0.84    &\ \ 0.86   \\
		BERT-MT   & 0.91       & 0.87   &\ \ 0.89   \\
		\noalign{\hrule height 0.5pt}
		Our Method   & 0.94       & 0.92    &\ \ 0.93   \\
		Ensemble    & \textbf{0.97}      & \textbf{0.94}    & \ \ \textbf{0.96}  \\	
		\noalign{\hrule height 1pt}
	\end{tabular}
	\label{sci}
\end{table*}
\subsection{Pre-training strategies}
We use the continuing pre-training strategy with the proposed method using the Sci-BERT model\footnote{https://huggingface.co/allenai/scibert\_scivocab\_cased}. 
Except for the dataset of the competition (SDU@AAAI-22), we additionally used the dataset of the SDU@AAAI-21 and the science paper data set\footnote{https://huggingface.co/datasets/scientific\_papers} for continuing pre-training. The ratio of the positive sample and negative sample keeps 1:2, where the negative sample can be obtained from different short-form acronyms. 
To obtain good sentence representations, we add the phrase-level contrastive pre-training objective into the pre-training for 200k steps, another 200k is to perform the original BERT pre-training tasks. 
The training samples are truncated with a maximum length of 300 and the batch size is set as 32. The temperature $\tau$ in Eqn. \ref{eq1} is set to 2e-2. For optimization, we use the same AdamW optimizer \cite{loshchilov2017decoupled} with weighted decay. The initial learning rate is 1e-4 for warmups about 10\% of the total steps. We implement the pre-training step with 8 NVIDIA 3090 GPUs with 24GB memory.
\subsection{Implementation}
As for the RoBERTa and the SciBERT model, we fine-tune with initial learning rates of the 5e-5 optimizing via the AdamW optimizer with batch size of 32. 
\par
As for the hdBERT and the BERT-MT model, we follow the default setting the same as the paper \cite{pan2021bert}, where the RoBERTa\footnote{https://huggingface.co/roberta-large} and the
SciBERT\footnote{https://huggingface.co/allenai/scibert\_scivocab$\_$uncased} is adopted from the Transformers of the Huggingface \cite{wolf2019huggingface}.
\par
The BERT-MT model is fine-tuned for 15 epochs with a batch size of 32. The initial learning rate for the encoder is 1e-5, and the others are 5e-4. The minimum learning rate is 5e-7 with the 
Adam optimizer \cite{kingma2014adam}.
\par
Our fine-tuning stage is implemented with a batch size of 32 for 15 epochs. We utilize the trained student model for fine-tuning the test experiments.
The AdamW optimizer is adopted with an initial learning rate of 1e-4 and annealed gradually after a warm-up epoch until it reached 1e-5. The weight hyper-parameter $\lambda$ is set to 0.5 to accelerate the whole training stage.
\par
As for the ensemble part, we fuse the output probability of the different baselines and add the balanced weights to get the final predictions, where more implemented details can be found in the work \cite{fung2006balanced}.
\begin{table}[]
	\centering
	\renewcommand\arraystretch{1.2}
	\setlength{\tabcolsep}{5mm}
	\caption{Online leaderboard.}
	\begin{tabular}{cccc}
		\noalign{\hrule height 1pt}
		Method     & Precision  & Recall  & \ \ \ F1 \\ \noalign{\hrule height 0.5pt}
		Rank1          & \textbf{0.97}      &  \textbf{0.94}    &   \ \ \textbf{0.96}      \\ 
		Rank2                 & 0.95  &  0.90  &  \ \ 0.93   \\ 
		Rank3             & 0.88  &   0.82  &  \ \   0.85    \\
		Rank4                  & 0.81   &    0.77  & \ \  0.79    \\ 
		Rank5               & 0.81   &  0.69     &   \ \ 0.75   \\ 
		Baseline              & 0.74  &   0.37     &\ \ 0.49     \\ 
		\noalign{\hrule height 1pt}
	\end{tabular}
	\label{onl}
\end{table}
\section{Results}
The main results of our model and baselines are shown
in Table \ref{sci}. It can be found that the performance of the pre-trained model outperforms the rule-based method since the rule-based method is difficult to pick the correct phrase from confusing acronym options from the dictionary due to its poor generalization. The SciBERT beats the RoBERTa in the three scores, which indicates that the domain-specific pre-training is of significant for science document understanding. The scientific domain pre-trained model can capture a deep representation of the confusing acronyms. The hdBERT merges different types of hidden features to get better generalization in binary classification, thereby performing well in this task. The results of the BERT-MT demonstrate that there are indeed many useful tricks in helping the model enhance the ability of robustness. It is noted that the proposed method outperforms the other baselines in three scores, which represents that the pre-trained model with continuing contrastive pre-training can further improve the model's ability to represent acronyms. Notice that the ensemble method can further improve the diversity of the final results thereby achieving the best performance in the test set. In summary, we finally rank the 1-st in the online leaderboard, which is shown in Table \ref{onl}.
\section{Conclusion}
We describe a simple framework for contrastive learning of acronym disambiguation in the shared task 2 of the SDU@AAAI-22.
Many baselines are implemented to compare with the proposed method, including methods based on pre-training, combinations of different structures, and useful tricks. The results demonstrate that the proposed method outperforms all other baselines, achieving the best performance (top-1) in the acronym disambiguation of scientific English. It can be further concluded that the continuing contrastive pre-training method can enhance the model's ability to represent the confusing phrases of the long-form acronym. The contrastive fine-tune can further enhance the generalization ability. In future work, we will extend our work as follows: (1) to use twin networks for training the teacher and the student together. (2) Adopting the fine-grained and the coarse-grained embedding into the contrastive pre-training to better acknowledge the meaning of the sentence.

\section*{Acknowledgement}
{We would like to thank Yixuan Su and all anonymous review experts for their valuable suggestions to improve the paper. This work is supported by the National Key Research and Development Project of China (2018YFB1305200) and the National Natural Science Fund of China (62171183, 61801178).}

\bibliography{references}

\end{document}